\documentclass[11pt]{article}
\RequirePackage[OT1]{fontenc}
\RequirePackage{amsthm,amsmath,natbib}
\RequirePackage[colorlinks]{hyperref}
\usepackage{amsfonts}
\usepackage{amssymb}
\usepackage{amstext}
\usepackage{parskip}
\usepackage{fullpage}
\usepackage{amssymb}

\bibpunct{(}{)}{;}{a}{,}{,}

\usepackage{times}

\usepackage{graphicx}
\usepackage{epsfig}
\usepackage{graphicx}
\usepackage{epstopdf}
\usepackage[footnotesize]{subfigure}
\usepackage{natbib}

\theoremstyle{plain}

\newcommand{\beq}{\begin{equation}}
\newcommand{\eeq}{\end{equation}}

\newcommand{\bes}{\begin{split}}
\newcommand{\ees}{\end{split}}

\newcommand{\benum}{\begin{enumerate}}
\newcommand{\eenum}{\end{enumerate}}

\newcommand\R{\mathbb{R}}

\newcommand{\mf}{\mathbf}

\newcommand{\e}{\mathbf{e}}

\newcommand{\w}{\mathbf{w}}
\newcommand{\x}{\mathbf{x}}
\newcommand{\y}{\mathbf{y}}

\newcommand{\cL}{{\cal L}}

\newcommand{\cN}{{\cal N}}

\newcommand{\E}{\mathbf{E}}

\newcommand{\myref}[1]{(\ref{#1})}

\newcounter{exampleI}
\newcounter{exampleII}
\newcounter{exampleIII}
\newcommand {\commentout}[1] {}

\DeclareFixedFont{\auacc}{OT1}{phv}{m}{n}{12}   

\def\math{${}^\diamond$}
\def\stat{${}^\star$}

\begin{document}

\title{Multi-task Sparse Structure Learning
}

\author{Andr\'e R Gon\c{c}alves\stat\math
 \;Puja Das\stat\; Soumyadeep Chatterjee\stat\; \\Vidyashankar Sivakumar\stat\; Fernando J Von Zuben\math\;
 Arindam Banerjee\stat\;\\ emails\footnote{andre@cs.umn.edu, pdas@cs.umn.edu, chatter@cs.umn.edu, sivakuma@cs.umn.edu, vonzuben@dca.fee.unicamp.br,
 \mbox{banerjee@cs.umn.edu}} \\ {\math}School of Electrical and Computer Enginnering, University of Campinas, Brazil\\
{\stat}Department of Computer Science, University of Minnesota, Twin Cities}

\maketitle
\begin{abstract}
Multi-task learning (MTL) aims to improve generalization performance by learning multiple related tasks simultaneously. While sometimes the underlying task relationship structure  is known, often the structure needs to be estimated from data at hand. In this paper, we present a novel family of models for MTL, applicable to regression and classification problems, capable of learning the structure of task relationships.
In particular, we consider a joint estimation problem of the task relationship structure and the individual task parameters, which is solved using alternating minimization. The task relationship structure learning component builds on recent advances in structure learning of Gaussian graphical models based on sparse estimators of the precision (inverse covariance) matrix.
We illustrate the effectiveness of the proposed model on a variety of synthetic and benchmark datasets for regression and classification. We also consider the problem of combining climate model outputs for better projections of future climate, with focus on temperature in South America, and show that the proposed model outperforms several existing methods for the problem.
\end{abstract}


\section{Introduction}
\label{sec:introduction}

The fundamental assertion of multi-task learning (MTL) is that learning over multiple related tasks can outperform
learning each task in isolation. The past few years have seen an increase in activity in this area where new methods
and applications have been proposed. From the methods perspective, there have been contributions to novel formulations
for describing task structure and incorporating them in the learning framework \cite{Evgeniou2004,jrsr10,jiye09,Kim2010,Obozinski2010}.
Meanwhile MTL has been applied to problems ranging from object detection in computer vision to web image and video
search \cite{Wang2009} to multiple microarray data set integration in computational biology \cite{Kim2010,Widmer2012}, to
name a few.

Much of the existing work in MTL assumes the tasks relationship
structure to be known (see Section \ref{sec:related_work}). However, in many problems, there is only a high level understanding
of the task relationships, and hence the structure of the tasks relationship needs to be learned from the data.


Considerable recent advances have been made in the area of structure learning in probabilistic graphical models, where one estimates the (conditional) dependence structure
between random variables in a high-dimensional distribution~\cite{laur96, mebu06, baed08,Hsieh2012,Wang2013}. In particular, assuming sparsity in the conditional
dependence structure, i.e., each variable is dependent only on a few others, there are estimators
based on convex (sparse) optimization which are guaranteed to recover the correct dependency structure
with high probability, even when the number of samples is small compared to the number of variables.


In this paper, we present a family of models for MTL, collectively referred to as Multi-task Sparse Structure Learning (\!MSSL\!), capable of learning the structure of task
relationships as well as parameters for individual tasks. For MSSL, we consider a joint estimation problem of the task relationship structure and the individual task parameters, which is solved using alternating minimization. The task relationship structure learning component builds on recent advances in structure learning of Gaussian graphical models based on sparse estimators of the precision (inverse convariance) matrix~\cite{mebu06, baed08,Hsieh2012,Wang2013}.

MSSL has important practical implications:
given a set of tasks, one can just feed the data from all the tasks without any knowledge or guidance on
tasks relationship structure, and MSSL will figure out which tasks are related and also estimate task specific parameters.
The structure is learned by considering a multivariate Gaussian prior with sparse precision (inverse covariance) matrix, so that the structure learning problem in the multi-task learning setting reduces to the structure learning problem for a Gaussian graphical model. We consider two different ways of imposing the sparse precision structure---first, across the task specific parameters for regression or classification, and second, across the task specific residual errors for regression.
The corresponding estimation/optimization problems
are solved using suitable first order methods, including proximal updates
and alternating direction method of multipliers.


MSSL can be applied to regression and classification problems, and we perform experiments on both classes of problems.  Through experiments on a wide variety of datasets, we illustrate that MSSL is competitive with and usually outperforms several baselines from the existing
MTL literature. Furthermore, the task relationships learned by MSSL are accurate and
consistent with domain knowledge on the problem.
We also consider the problem of combining climate model outputs for better projections of future climate,
with focus on temperature in South America. With temperature forecast in each location as a task, MSSL estimates a meaningful dependency structure, e.g., connecting places around the Amazon, but not connecting places across the Andes, etc., without any knowledge of the spatial locations of the tasks, and outperforms several existing methods for the problem.

The rest of the paper is structured as follows. Section~\ref{sec:related_work} briefly discusses related work in
multi-task learning. Section~\ref{sec:multitask_struct_learn} gives a gentle
introduction to the proposed multi-task sparse structure learning (MSSL) approach. Section~\ref{sec:parameter_structure} discusses a
specific form of MSSL where the task dependency structure is learned based on the task parameters,  which we refer to
as $p$-MSSL. Section~\ref{sec:residual_structure} discusses another specific form of MSSL where the task
dependency structure is learned based on the task residual errors, which we refer to as $r$-MSSL.
Section~\ref{sec:experiments} discusses experimental results on regression and classification using
synthetic, benchmark, and climate datasets. We conclude in Section~\ref{sec:conclusion}.


\section{Related Work}
\label{sec:related_work}

One of the key questions in MTL is to identify which tasks are related to each other, so that
it can incorporate this knowledge when learning all tasks simultaneously. Based on the relationship
assumed between the tasks we can broadly classify MTL methods into three classes:
(1) all tasks are related, (2) tasks are organized in clusters, and (3) tasks are related in a tree/graph structure.

The first class of MTL methods assume that all tasks are related and information about a task is
shared with all other tasks. \cite{Evgeniou2004} assumes all task parameters are close to each other.
\cite{Obozinski2010} constrains all tasks to share a common set of features.
\cite{jrsr10, jiye09} make sparsity and low-rank assumptions about the structure of the parameter matrix.

In clustered multi-task learning, the hypothesis is that models of tasks from the same group are
closer to each other compared to tasks from a different group. In \cite{Bakker2003} task
clustering is enforced by considering a mixture of Gaussians as a prior over task coefficients.
\cite{evmp05} proposes a task clustering regularization to encode cluster information in the MTL formulation.
\cite{Xue2007} employs a non-parametric prior distribution over task coefficients to encourage
task clustering.

For graph structured MTLs, two tasks are related if they are connected in a graph, i.e. the
connected tasks are similar. The similarity of two related tasks can be represented by the
weight of the connecting edge~\cite{Kim2010,zhou2011a}.


Our focus in this work is on problems which model task relatedness as an undirected graph. We learn such a
structure assuming it is not available a priori. The model proposed in \cite{Zhang2012} is closest to our
work. In \cite{Zhang2012} a matrix-variate normal distribution is used as a prior for the matrix
representing the model parameters for all tasks, and then such prior distribution hyperparameter captures
the covariance matrix among all task coefficients. While their model learns the covariance matrix, we
directly learn the (sparse) {\it precision} matrix (inverse of covariance matrix) among tasks parameters,
which has a meaningful interpretation in terms of conditional independence and partial correlation. To the
best of our knowledge, this is the first MTL approach which measures the relationship among tasks by
using partial correlation and also learns it from the data.



\section{Multi-task Structure Learning}
\label{sec:multitask_struct_learn}

In this section we describe the basic idea of our framework. We start by giving a small description of
the notations used throughout the paper, followed by a short introduction to the structure estimation problem.

\subsection{Notation and Preliminaries}
Let $K$ be the number of tasks, $d$ be the number of covariates which are shared across all tasks, and
$n_k$ be the number of samples for each task. Also $\mf X_k\in \mathbb{R}^{n_k\times d}$ be the covariates
matrix (input data) for the $k$-th task, $\mf y_k\in \mathbb{R}^{n_k\times 1}$ be the response
for the $k$-th task, and $\mf W\in \mathbb{R}^{d\times K}$ be the parameter matrix,
where columns are parameters $\mathbf{w}_k$ for each task and $\mathbf{\hat{w}}_j$ denotes the $j$-th row of $\mf W$.

For regression, we consider a linear model $\mf y_k = \mf X_k \mf w_k + \mathbf{e}_k$ for each task
where $\mathbf{e}_k$ is the residual error for the $k$-th task. When we consider task relatedness
based on the residual errors $\mf e_k$, we assume that all tasks have the same number of training samples $n$
and we define the matrix $\mf E  = (\mf e_1,\mf e_2, ..., \mf e_K) \in \mathbb{R}^{n\times K}$.
We denote the $j$-th row of $\mf E$ as $\mathbf{\hat{e}}_j$. Also $\boldsymbol{\Omega} \in \mathbb{R}^{K\times K}$ is
a matrix that captures the task relationship structure.


MTL can benefit from knowing the underlying structure relating the tasks while learning parameters for each task.
In situations where some of the tasks may be highly dependent on each other, the strategy of isolating each
task does not exploit the potential information one may acquire from other related tasks. We first describe
how we capture the task relatedness and then we introduce our multi-task sparse structure
learning (MSSL) framework, which is data driven and learns the structure of task
relatedness while learning the parameters for each task.

\subsection{Structure Estimation}


Let $\mf r = (r_1,\ldots,r_d)$ be a $d$-variate random vector with joint distribution $\mathcal{P}$.
Any such distribution can be characterized by an undirected graph $G = (V,E)$, where the vertex set $V$
represents the $d$ covariates of $\mathbf{r}$ and edge set $E$ represents the conditional dependence
relations between the covariates of $\mf r$. If $r_i$ is conditionally independent of $r_j$ given the other variables,
then the edge $(i,j)$ is not in $E$. Assuming $\mathbf{r}\sim\mathcal{N}(\mathbf{0},\boldsymbol{\Sigma})$,
the missing edges correspond to zeros in the inverse covariance matrix or \textit{precision} matrix given
by $\boldsymbol{\Sigma}^{-1} = \boldsymbol{\Omega}$, i.e., $(\boldsymbol{\Sigma}^{-1})_{ij} =0 \ \forall (i,j) \notin E$.

Classical estimation approaches \cite{demp72} work well when $d$ is small. Given $n$ i.i.d. samples
$\mathbf{r}_1,\ldots,\mathbf{r}_n$ from the distribution, the empirical covariance matrix is
$\boldsymbol{\hat{\Sigma}} = \frac{1}{n} \sum_{i=1}^n (\mathbf{r}_i - \mathbf{\bar{r}})^T (\mathbf{r}_i -
\mathbf{\bar{r}})$, where $\mathbf{\bar{r}} = \frac{1}{n} \sum_{i=1}^n \mathbf{r}_i$.
However when $d \gg n$, $\boldsymbol{\hat{\Sigma}}$ is rank-deficient and its inverse cannot be used to estimate the precision
matrix $\boldsymbol{\Omega}$. However, for a sparse graph where most of the entries in the
precision matrix are zero, several methods exist that can estimate $\boldsymbol{\Omega}$~\cite{mebu06,frht08}.

\subsection{MSSL Formulation}

For ease of exposition, consider a simple linear model, $\y_k = \mf X_k \w_k + \mathbf{e}_k$, for each task.
MSSL will learn both the task parameters $\mathbf{w}_k$ for all tasks and the structure which will be
estimated based on some information from each task. Further, the structure is used as inductive
bias in the model parameter learning process so that it may improve tasks generalization capability.
So, the learning of one task is biased by other related tasks \cite{Caruana:1993}.




We investigate and formalize two ways of learning the relationship structure, represented
by $\boldsymbol{\Omega}$: (a) modeling $\boldsymbol{\Omega}$ from the task specific parameters
$\mathbf{w}_k, \forall k=1,..,K$, and (b)
modeling $\boldsymbol{\Omega}$ from the residual errors $\boldsymbol{e}_k, \forall k=1,..,K$. Based on how
we model $\boldsymbol{\Omega}$, we propose $p$-MSSL (from tasks parameters) and $r$-MSSL
(from residual error). 




At a high level, the estimation problem in MSSL is of the form:
\beq
\min_{\mf W,\boldsymbol{\Omega}\succ 0}~ \sum_{k=1}^{K}\mathcal{L}((\mf y_k,\mf X_k),\mf w_k) + \mathcal{B}(\mf W,\boldsymbol{\Omega}) + \mathcal{R}_1(\mf W) + \mathcal{R}_2(\boldsymbol{\Omega}),
\label{eq:general_mssl}
\eeq
where $\mathcal{L}(\cdot)$ denotes a suitable task specific loss function, $\mathcal{B}(\cdot)$ is the
inductive bias term, and $\mathcal{R}_1(\cdot), \mathcal{R}_2(\cdot)$ denote suitable sparsity
inducing regularization terms. The interaction between parameters $\mathbf{w}_k$ and
the relationship matrix $\boldsymbol{\Omega}$ is captured by the $\mathcal{B}(\cdot)$ term. When
$\boldsymbol{\Omega}_{k,k'} = 0$, the parameters $\w_k$ and $\w_{k'}$ have no influence on each other.

\section{Parameter Precision Structure}
\label{sec:parameter_structure}



If the tasks are unrelated, one can learn the columns $\mf w_k$ of the parameter matrix $\mf W$ independently for each of the $K$ tasks. However, if there exist relationships between the $K$ tasks, learning the columns of $\mf W$ independently fails to take advantage of these dependencies. In such a scenario, we propose to utilize the precision matrix $\boldsymbol{\Omega} \in \R^{K\times K}$ among the tasks in order to capture pairwise partial correlations.

In the $p$-MSSL model we assume that each \emph{row} $\hat{\w}_j$ of the matrix $\mf W$ follows a
multivariate Gaussian distribution with zero mean and covariance matrix $\boldsymbol{\Sigma}$, i.e.
$\hat{\w}_j \sim \cN(\mathbf{0},\boldsymbol{\Sigma})~\forall j=1,\ldots,d$, where
$\boldsymbol{\Sigma}^{-1}$=$\boldsymbol{\Omega}$. The problem then is to estimate both
the parameters $\w_{1},\ldots,\w_K$ and the precision matrix $\boldsymbol{\Omega}$.
By imposing such a prior over the {\em rows} of $\mf W$, we are capable of explicitly estimating the dependency structure between
the tasks via the precision matrix $\boldsymbol{\Omega}$.

Having a multivariate Gaussian prior over the {\em rows} of $\mf W$, its posterior can be written as
\small
\beq
P\left(\mf W\middle|(\mf X,\mf Y), \boldsymbol{\Omega}\right) \propto \prod_{k=1}^{K}\prod_{i=1}^{n_{k}}P\left(y_{k}^{(i)}\middle|\x_{k}^{(i)}, \w_{k} \right)\prod_{j = 1}^{d}P\left(\hat{\w}_{j}| \boldsymbol{\Omega}\right)~,
\label{eq:likelihood}
\eeq
\normalsize
where the first term in the right hand side denotes the conditional distribution of the response given the input and parameters, and the second term denotes the prior over {\em rows} of $\mf W$. In this paper, we consider the \emph{penalized} maximization of~\eqref{eq:likelihood}, assuming that the parameter matrix $\mf W$ and the precision matrix $\boldsymbol{\Omega}$ are sparse. We provide two specific instantiations of this model. First, we consider a Gaussian conditional distribution, wherein we obtain the well known least squares regression problem (Section~\ref{subsec:least_squares}). Second, for discrete labeled data, choosing a Bernoulli conditional distribution leads to a logistic regression problem (Section~\ref{subsec:logregression}).

\subsection{Least Squares Regression}
\label{subsec:least_squares}
Assume that
\beq
P\left(y_{k}^{(i)}\middle|\x_{k}^{(i)}, \w_{k}\right) = \cN\left(y_k^{(i)}\middle|\w_k^T\x_k^{(i)},\sigma\right)~,
\eeq
where we assume $\sigma=1$ and consider penalized maximization of~\eqref{eq:likelihood}. We can write this optimization problem as minimization of the negative logarithm of~\eqref{eq:likelihood}, which corresponds to a linear regression problem~\cite{hatr01}
\small
\begin{equation}
\min\limits_{\mf W, \boldsymbol{\Omega}\succ 0} \left\{\frac{1}{2}\sum_{k = 1}^{K}\|\mathbf{X}_k\mathbf{w}_k-\mathbf{y}_k\|^2_2- \frac{K}{2}\log|\boldsymbol{\Omega}|
+ \text{Tr}(\mf W\boldsymbol{\Omega}\mf W^{T})\right\}~.
\end{equation}
\normalsize

Further, assuming $\boldsymbol{\Omega}$ and $\mf W$ are sparse, we can add $\ell_1$-norm regularizers over both parameters and obtain a regularized regression problem~\cite{hatr01}
%
%
%
\begin{equation}
\min\limits_{\mf W, \boldsymbol{\Omega}\succ 0} \left\{ \frac{1}{2}\sum_{k = 1}^{K}\|\mathbf{X}_k\mathbf{w}_k -\mathbf{y}_k\|^2_2 -\frac{K}{2}\log|\boldsymbol{\Omega}| + \text{Tr}(\mf W\boldsymbol{\Omega}\mf W^{T})  + \lambda \|\boldsymbol{\Omega}\|_{1}+\gamma\|\mf W\|_1\right\}~,
\label{eq:regression_obj}
\end{equation}
where $\lambda,\gamma>0$ are penalty parameters.

In this formulation, the term involving the trace outer product $\text{Tr}(\mf W\boldsymbol{\Omega}\mf W^{T}) $ affects the \emph{rows} of $\mf W$, such that if $\boldsymbol{\Omega}_{ij}\neq 0$, then the columns $\w_i$ and $\w_j$ are constrained to be similar. 
We also observe that for a fixed $\boldsymbol{\Omega}$, since we have the inductive bias term, the regularization over $\mathbf{W}$ is more like the elastic-net penalty, since it has both $\ell_{1}$ and $\ell_{2}$
regularization on $\mathbf{W}$. Such elastic-net type penalties have the added advantage of picking up correlated
relevant covariates (unlike Lasso), but have the downside that they are hard to interpret statistically.
The objective function is nevertheless well defined from an optimization perspective. Also we note that modeling the relationship as a multivariate Gaussian distribution is not necessarily a restrictive assumption. Recently there has been work in developing the Gaussian copula family of models~\cite{lhyfw12} which addresses a large class of problems commonly encountered. Our framework can be extended to such models.

Next, we illustrate methods to solve the optimization problem~(\ref{eq:regression_obj}) efficiently. We propose an iterative optimization algorithm which alternates between updating $\mf W$ (with fixed $\boldsymbol{\Omega}$) and $\boldsymbol{\Omega}$ (with fixed $\mf W$). This results in alternating between solving two non-smooth convex optimization problems in each iteration.

\noindent \textbf{Alternating Optimization: }
The alternating minimization algorithm proceeds as follows.
\begin{enumerate}
\item Initialize $\boldsymbol{\Omega}^0=I_K$ and $\mf W^0=0_{d\times K}$
\item For $t=1,2,\ldots$ {\bf do}
\begin{equation}
   \mf W^{(t+1)}|\boldsymbol{\Omega}^{(t)} = \underset{\mf W}{\text{argmin}} \left\{\frac{1}{2}\sum_{k=1}^{K}\|\mathbf{X_k}\mathbf{w}_k-\mathbf{y}_k\|_2^2+ \text{Tr}(\mf W\boldsymbol{\Omega}^{(t)} \mf W^T) +\gamma\|\mf W\|_1 \right\} \label{eq:w_sub_1}
\end{equation}
\begin{equation}
   \boldsymbol{\Omega}^{(t+1)}|\mf W^{(t+1)} = \underset{\boldsymbol{\Omega} \succ 0}{\text{argmin}}\left\{ \text{Tr}(\mf W^{(t+1)}\boldsymbol{\Omega} \mf W^{T(t+1)})-\frac{K}{2}\log \left| \boldsymbol{\Omega}\right| +  \lambda \lVert \boldsymbol{\Omega} \rVert _1 \right\}  \label{eq:sig_sub_2}
\end{equation}
\end{enumerate}

Note that this alternating minimization procedure is guaranteed to converge to a minima, since the original problem~\eqref{eq:regression_obj} is convex in each argument $\boldsymbol{\Omega}$ and $\mf W$~\cite{guby05}.


\noindent{\textbf{Update for $\mf W$:}}
The update step~\eqref{eq:w_sub_1} is a general case of the formulation proposed by~\cite{Karthik2013}
in the context of climate model combination, where in our proposal $\boldsymbol{\Omega}$ is any positive
definite precision matrix, rather than a fixed Laplacian matrix.



Using $\operatorname{vec}()$ notation we can re-write the optimization problem in~\eqref{eq:w_sub_1} as
\begin{equation}
  \min_{\mf W} \left\{\frac{1}{2} \|\bar{\mf X}\operatorname{vec}(\mf W) - \operatorname{vec}(\mf Y)\|_2^2  + \frac{1}{2}\operatorname{vec}(\mf W)^T\mathbf{P}(\boldsymbol{\Omega} \otimes I_d)\mathbf{P}^T\operatorname{vec}(\mf W) + \gamma \|\mf W\|_1\right\}
\end{equation}

\noindent where $\otimes$ is the Kronecker product, $I_d$ is a $d\times d$ identity matrix,
and $\mathbf{P}$ is a permutation matrix that converts the column stacked arrangement of $\operatorname{\mf W}$
to a row stacked arrangement. $\mathbf{\bar{X}}$ is a block diagonal matrix where the main
diagonal blocks are the task data matrices $\mathbf{X}_k, \forall k=1,..,K$ and the off-diagonal blocks are zero matrices.
Therefore the problem is a $\ell_1$ penalized quadratic optimization program, which we solve using
established proximal gradient descent methods such as FISTA~\cite{bete09}.


We note here that the optimization of~\eqref{eq:w_sub_1} can be scaled up considerably by using an
ADMM~\cite{Boyd2011} based algorithm, which decouples the non-smooth $\ell_1$ term from the smooth convex
terms in the objective of~\eqref{eq:w_sub_1}. Alternating method of multipliers (ADMM) \cite{Boyd2011} is
a strategy that is intended to blend the benefits of dual decomposition and augmented Lagrangian methods
for constrained optimization. It takes the form of a {\it decomposition-coordination} procedure, in which
the solutions to small local problems are coordinated to find a solution to a large global problem \cite{Boyd2011}.

%
%
%
%
%
%

\noindent{\textbf{Update for $\boldsymbol{\Omega}$:}}
The update step for $\boldsymbol{\Omega}$, given in~\eqref{eq:sig_sub_2}, is known as
{\it sparse inverse covariance selection problem} \cite{baed08} which can be efficiently solved using ADMM~\cite{Boyd2011}.
We refer interested readers to~\cite{Boyd2011} Section 6.5 for details on derivation of the updates.
Given the matrix $\mf W^{(t+1)}$, \eqref{eq:sig_sub_2} can be solved to obtain $\boldsymbol{\Omega}^{(t+1)}$ by iterating the following ADMM steps.
\begin{enumerate}
\item Initialize $\boldsymbol{\Theta}^{0}=\boldsymbol{\Omega}^{(t)}$, $\mf Z^0=0_{K\times K}$, $\mf U^0=0_{K\times K}$.
\item For $l=1,2,\ldots$ {\bf do}
 \begin{equation}
          \boldsymbol{\Theta}^{(l+1)} := \underset{\boldsymbol{\Theta} \succ 0}{\text{argmin}} \left\{\text{Tr}(\mf W^{(t+1)}\boldsymbol{\Theta} {\mf W^{(t+1)}}^T)-\frac{K}{2}\log|\boldsymbol{\Theta}|  +\frac{\rho}{2} \|\boldsymbol{\Theta}-\mf Z^{(l)}+\mf U^{(l)}\|^2_F \right\}
		  \label{eq:admm_1}
      \end{equation}
      \begin{equation}
          \mf Z^{(l+1)} := \underset{\mf Z}{\text{argmin}} \left\{\lambda\|\mf Z\|_1 + \frac{\rho}{2}\|\boldsymbol{\Theta}^{(l+1)}-\mf Z+\mf U^{(l)}\|^2_F \right\}
		  \label{eq:admm_2}
       \end{equation}
       \begin{equation}
          \mf U^{(l+1)} := \mf U^{(l)} + \boldsymbol{\Theta}^{(l+1)} - \mf Z^{(l+1)}
		  \label{eq:admm_3}
       \end{equation}
\item Output $\boldsymbol{\Omega}^{(t+1)} = \boldsymbol{\Theta}^l$, where $l$ is the number of steps for convergence.

\end{enumerate}

Each ADMM step can be solved efficiently. For the $\boldsymbol{\Theta}$-update, we can observe from the first order optimality condition of~\myref{eq:admm_1} and the implicit constraint $\boldsymbol{\Theta} \succ 0$ that
\begin{equation}
\rho \boldsymbol{\Theta}-\boldsymbol{\Theta}^{-1}=\rho(\mf Z^{(l)}-\mf U^{(l)})-\mf S,
\label{eq:inter_deriv}
\end{equation}
where $\mf S= {\mf W^{(t+1)}}^T\mf W^{(t+1)}$. Next we take the eigenvalue decomposition of $\rho(\mf Z^{(l)}-\mf U^{(l)})-\mf S=\mf Q\mf \Lambda \mf Q^T$, where $\mf \Lambda =\text{diag}(\lambda_1,\cdots,\lambda_K)$ and $\mf Q^{T}\mf Q=\mf Q\mf Q^T=I$. We now multiply~\myref{eq:inter_deriv} by $ \mf Q^T$ on the left and by $\mf Q$ on the right to get $\rho\widetilde{\boldsymbol{\Theta}}-\widetilde{\boldsymbol{\Theta}}^{-1}=\Lambda$, where $\widetilde{\boldsymbol{\Theta}}=\mf Q^{T}\boldsymbol{\Theta} \mf Q$.

\begin{equation}
{\widetilde{\boldsymbol{\Theta}}}_{ii} = \dfrac{\lambda_i+\sqrt{\lambda_i^2+4\rho}}{2\rho}
\end{equation}
Now $\boldsymbol{\Theta}= \mf Q \widetilde{\boldsymbol{\Theta}} \mf Q^{T}$ satisfies~\myref{eq:inter_deriv}.

The $\mf Z$-update~\myref{eq:admm_2} can be computed in closed form:
\beq
\mf Z^{(l+1)}=S_{\lambda/\rho}\Big(\boldsymbol{\Theta}^{(l+1)}+\mf U^{(l)}\Big),
\eeq
where $S_{\lambda/\rho}(\cdot)$ is an element-wise soft-thresholding operator~\cite{Boyd2011}.


\subsection{Log Linear Models}
\label{subsec:logregression}

As described previously our model can also be applied to classification. Let us assume that
\beq
P\left(y_k^{(i)}\middle|\x_k^{(i)},\w_k\right) = \operatorname{Be}\left(y_k^{(i)}\middle|h\left(\mathbf{w}_k^T\mathbf{x}_k^{(i)}\right)\right)~,
\label{eq:glm conditional}
\eeq
where $h(\cdot)$ is the sigmoid function, and  $\operatorname{Be}(p)$ is a Bernoulli distribution.
Therefore, following the same construction as in Section~\ref{subsec:least_squares}, learning the parameters $\mf W$, and $\boldsymbol{\Omega}$ can be achieved by optimization of the following objective.

\begin{equation}
\min\limits_{\mathbf{\mf W}, \boldsymbol{\Omega}\succ 0} \left\{\frac{1}{2}\sum_{k=1}^{K}\sum_{i=1}^{n_{k}}\Big\{y_{k}^{(i)}\w_{k}^{T}\x_{k}^{(i)} - \log\Big(1+e^{\w_{k}^{T}\x_{k}^{(i)}}\Big)\Big\} -\frac{K}{2}\log |\boldsymbol{\Omega}| + \text{Tr}(\mathbf{W}\boldsymbol{\Omega}\mathbf{W}^{T}) + \lambda\|\boldsymbol{\Omega}\|_{1}+\gamma\|\mathbf{W}\|_{1}\right\}~.
 \label{eq:glm objective}
\end{equation}

The loss function in Eq.~(\ref{eq:glm objective}) is the logistic loss, where we have considered a 2-class classification setting.
Note that the objective function is similar to the one obtained for multi-task
learning with linear regression in Section~\ref{subsec:least_squares}. Therefore, we use the same
alternating minimization algorithm described in Section~\ref{subsec:least_squares} to optimize \eqref{eq:glm objective}.

In general, we can consider any Generalized Linear Model (GLM)~\cite{neld72}, with different link functions $h(\cdot)$, and therefore different probability density functions such as Poisson, Gamma etc. for the conditional distribution~\eqref{eq:glm conditional}. For any such model, our framework requires optimization of an objective of the form
\begin{equation}
\min_{\mf W,\boldsymbol{\Omega}\succ 0}\left\{\sum_{k=1}^{K}\cL(\y_k,\mf X_k\mathbf{w}_k)+\frac{1}{2}\text{Tr}(\mf W\boldsymbol{\Omega} \mf W^T)
			 - \frac{K}{2} \log \left|\boldsymbol{\Omega}\right| + \lambda \|\boldsymbol{\Omega}\|_1 + \gamma\|\mathbf{W}\|_1 \right\}
\label{eq:prob_min}
\end{equation}
where $\cL(\cdot)$ is a convex loss function obtained from a GLM.

\section{Residual Precision Structure}
\label{sec:residual_structure}


In $r$-MSSL we assume that the rows of the residual error matrix $\mf E$,
$\mathbf{\hat{e}}_j \sim \mathcal{N}(0,\boldsymbol{\Sigma}), \forall j=1,...,n$, where
$\boldsymbol{\Sigma}^{-1}=\boldsymbol{\Omega}$, and $\boldsymbol{\Omega}$ is
the precision matrix among the tasks. In contrast to $p$-MSSL the relationship among tasks is modeled in terms
of partial correlations among the errors $\e_1,\ldots,\e_K$ instead of considering explicit dependencies
between the parameters $\w_1,\ldots,\w_K$ for the different tasks.


Finding the dependency structure among the tasks now amounts to estimating the precision matrix
$\boldsymbol{\Omega}$. Such models are commonly used in spatial statistics~\cite{mama84} in order to
capture spatial autocorrelation between geographical locations. For example,
in domains such as climate or remote sensing, there often exist noise autocorrelations over the spatial
domain under consideration. Incorporating this dependency through the precision matrix of the residual errors is
then more interpretable than explicitly modeling the dependency among the parameters $\mf W$.


We assume that the parameter matrix $\mf W$ is fixed, but unknown. Since the rows of $\mf E$ follow
a Gaussian distribution, maximizing the likelihood of the data, penalized with a sparse regularizer over
$\boldsymbol{\Omega}$, reduces to the following optimization problem
\beq
\min\limits_{\mf W, \boldsymbol{\Omega} \succ 0} \left\{  \sum_{k=1}^{K} \|\y_k- \mathbf{X}_{k}\w_{k} \|^2_2 -
\frac{K}{2} \log|\boldsymbol{\Omega}| + \text{Tr}(\E\boldsymbol{\Omega} \E^T) +  \lambda \|\boldsymbol{\Omega}\|_{1} + \gamma \|\mf W\|_1 \right\}~.
\label{eq:residual struc}
\eeq

We use the alternating minimization scheme illustrated in previous sections to optimize the above objective. Note that since the objective is convex in each of its arguments $\mf W$ and $\boldsymbol{\Omega}$, and thus is guaranteed to reach a local minima through alternating minimization~\cite{guby05}. Further, the model can be extended to losses other than the squared loss which we obtain by assuming that the columns of $\mf E$ are i.i.d Gaussian.


In the experiments section
$r$-MSSL was used for regression problems, while for classification we used $p$-MSSL.
In fact, $p$-MSSL can be applied to both regression and classification problems. $r$-MSSL, instead,
can only be applied to regression problems, as the residual error of a classification problem is
clearly not Gaussian.

\section{Experimental results}
\label{sec:experiments}

In this section we provide experimental results to show the effectiveness of the proposed framework
for both regression and classification problems.

\subsection{Regression}

For the regression problems we start with experiments on synthetic data and then move to the problem of
predicting air surface temperature in South America.

To select the penalty parameter $\lambda$ we used a stability selection procedure described
in \cite{Meinshausen2010}. It is a subsampling approach which provides a way to find stable structures
and hence a  principle to choose a proper amount of regularization for structure estimation.

\subsubsection{Synthetic dataset}

We created a synthetic dataset with 13 tasks. Each task has 100 data instances with 30 covariates so that we have
a data matrix in $\mathbb{R}^{100 \times 30}$. The training set for each task consists of 60 randomly chosen
rows from the data matrix so that we have $\mathbf{X} \in {\R}^{60 \times 30}$.
The parameter vector for each task $\w_{k}$ are chosen so that tasks 1-4 are related to each
other but unrelated to the remaining tasks. Similarly tasks 5-10 form
a cluster. Tasks 11, 12, and 13 are not related to any of the other tasks. In this way we have two clusters
of tasks and also three outlier tasks. The prediction variables are generated as
$\mf y_{k} = \mathbf{X}\w_{k} + \boldsymbol\xi$ where $\boldsymbol\xi \sim \mathcal{N}(0,I_{100})$. We train the
$p$-MSSL model on this dataset and test it on the remaining 40 data instances.

\begin{figure}[t]
\centering
\includegraphics[scale=0.45]{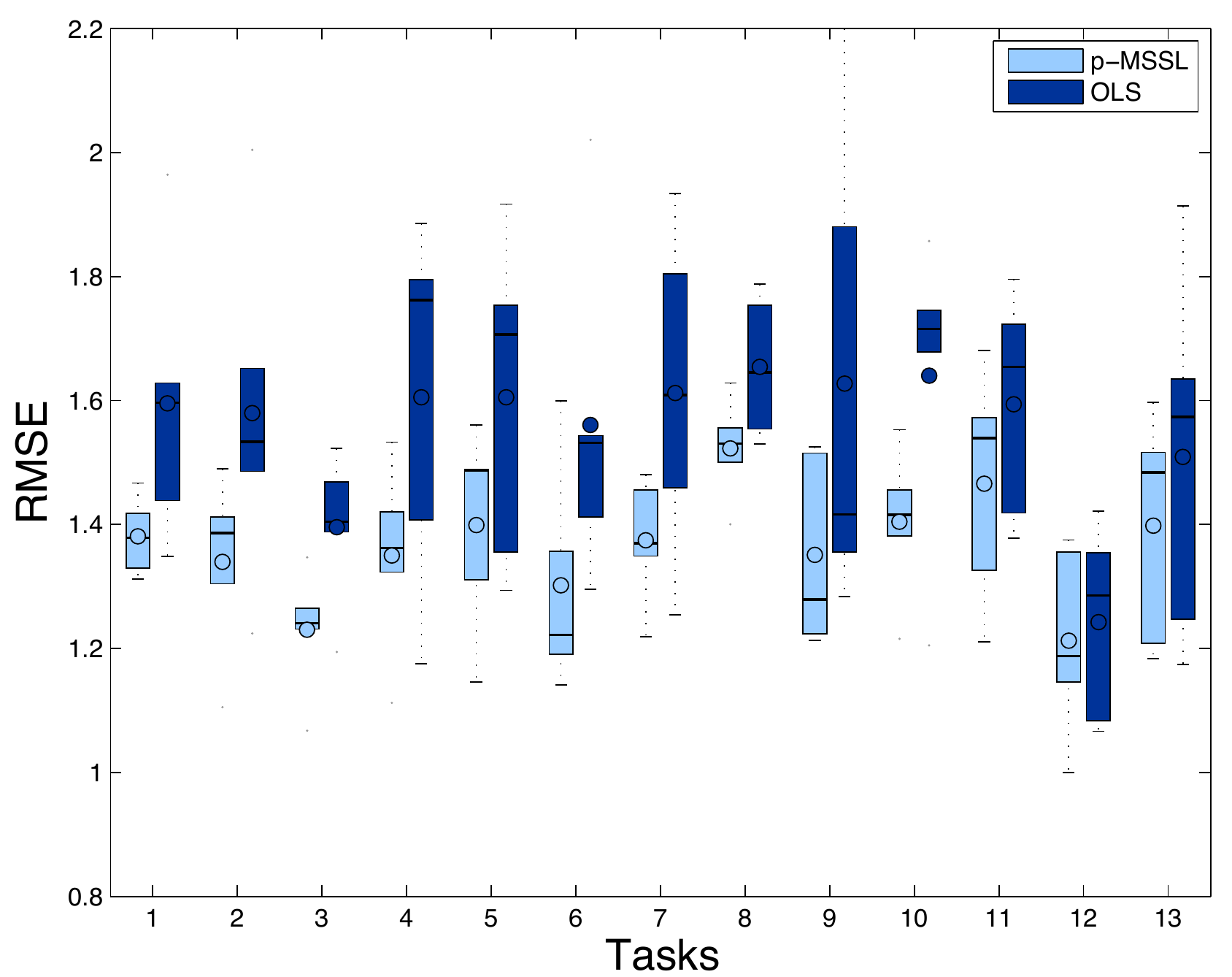}
\caption{RMSE error per task comparison between $p$-MSSL and Ordinary Least Square over 30 independent runs. $p$-MSSL
gives better performance on related tasks (1-4 and 5-10) but no significant improvement on unrelated tasks (11-13).}
\label{fig:boxplot_synth}
\end{figure}

Figure~\ref{fig:boxplot_synth} is a box-plot of RMSE error for $p$-MSSL and for the case
where Ordinary Least Squares (OLS) was applied individually for each task. As expected, sharing information
between tasks leads to better prediction accuracy. $p$-MSSL does well on related tasks 1-10 but there is
no significant improvement on tasks 11-13. Assuming a dense precision matrix or a precision matrix where tasks
11-13 are related to the other tasks leads to higher RMSE values.


\begin{figure}[hbt]
\centering
\hspace{-0.1cm}
\subfigure[]{\includegraphics[scale=0.30]{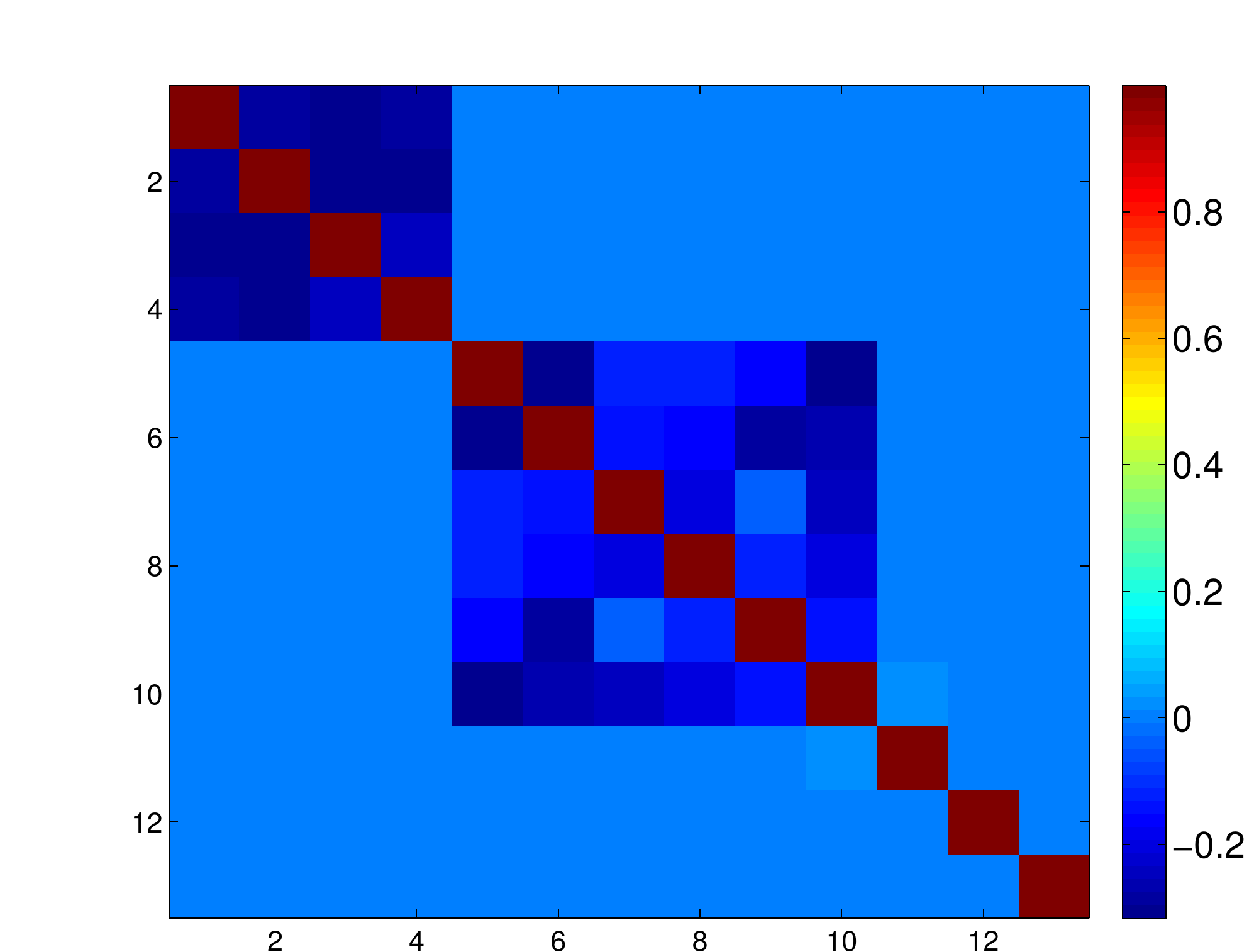}\label{fig:synth_mat_true}}
\subfigure[]{\includegraphics[scale=0.45]{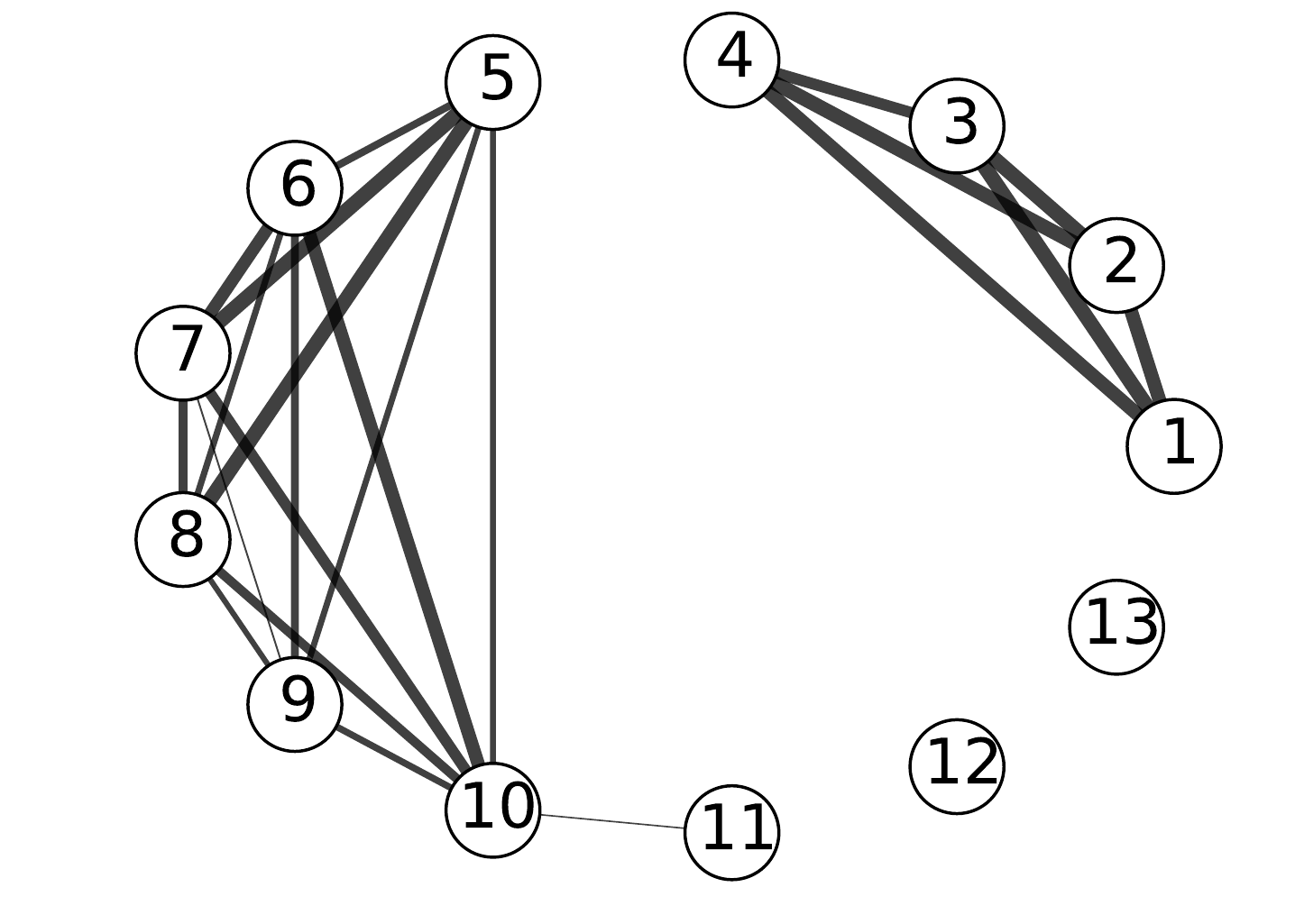}\label{fig:synth_mat_mssl}}
\caption{Structure estimated by the $p$-MSSL algorithm on the synthetic dataset: (a) precision matrix; (b) graph representation of the
task dependencies encoded by the precision matrix.}
\end{figure}

Figures~\ref{fig:synth_mat_true} and \ref{fig:synth_mat_mssl} depict the precision matrix estimated by $p$-MSSL
algorithm and its graph representation. The missing edges in the graph correspond to zeros in the precision
matrix, and it means that those pair of tasks are conditionally independent of each other given the other
tasks. As can be seen, our model is able to recover the true dependency structure among tasks.
The block structure is reflected in the estimated precision matrix.

\begin{figure}[htb]
\centering
\includegraphics[scale=0.45]{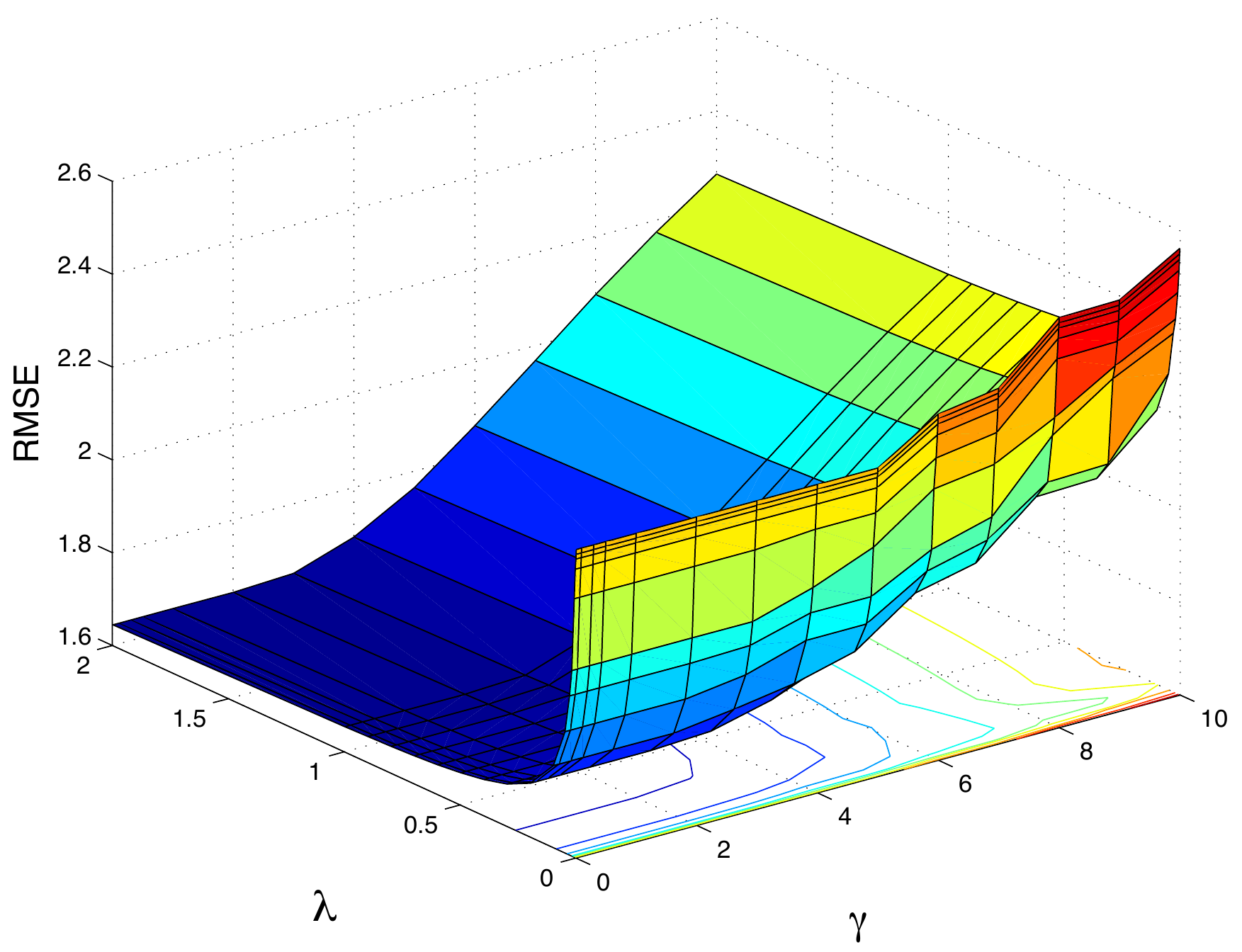}
\caption{Average RMSE error on the test set for all tasks varying parameters $\lambda$ (controls sparsity on $\boldsymbol{\Omega}$)
and $\gamma$ (controls sparsity on $\mf W$).}
\label{fig:sens_synth}
\end{figure}

Sensitivity analysis of $p$-MSSL sparsity parameters $\lambda$ (controls sparsity on $\boldsymbol{\Omega}$) and
$\gamma$ (controls sparsity on $\mf W$) on the synthetic data is presented in Figure~\ref{fig:sens_synth}.
As we increase $\gamma$ we encourage sparsity on $\mf W$ and having more zeros on $\mf W$ it becomes harder
for $p$-MSSL to capture the true relationship among the column vectors (tasks parameters), since it
learns $\boldsymbol{\Omega}$ based on $\bf W$. As a consequence of non-accurate structure estimation,
$p$-MSSL performance decreases as we can see in the figure. We also can see that for small $\lambda$ values $p$-MSSL
performance decreases, since we have a dense precision matrix which is different from the ground truth
(sparse matrix). In other words, learning an accurate precision matrix leads to an improvement of $p$-MSSL performance.



%
\subsubsection{Global Climate models combination}
\renewcommand{\arraystretch}{1.5}
\begin{table}[t]
\vspace{-4mm}
\centering
\caption{Global Climate Models description.}\footnotesize
\begin{tabular}{|l | l | c |}
\hline
{\bf GCM} & {\bf Origin} & {\bf Reference} \\
\hline
BCC\_CSM1.1	&	Beijing Climate Center, China & \cite{zwxd12} \\
CCSM4		&	National Center for Atmospheric Res., USA & \cite{wdba08} \\
CESM1		&	National Science Foundation, Department of Energy, NCAR, USA & \cite{smlb12} \\
CSIRO		&	Commonwealth Scientific and Industrial Research Organisation, Australia & \cite{csiro02} \\
HadGEM2	&	Met Office Hadley Centre, UK & \cite{hadgem11} \\
IPSL			&	Institut Pierre-Simon Laplace, France & \cite{dfdc12} \\
MIROC5		&	Atmosphere and Ocean Research Inst, Japan &  \cite{wsok10}\\
MPI-ESM		&	Max Planck Inst. for Meteorology, Germany & \cite{bbrg13} \\
MRI-CGCM3	&	Meteorological Research Inst., Japan & \cite{yuah12} \\
NorESM		&	 Norwegian Climate Centre, Norway & \cite{bbdi12} \\
\hline
\end{tabular}
\label{table:GCM_info}
\end{table}

A Global Climate Model (GCM) is a complex mathematical representation of the major climate system components (atmosphere, land surface, ocean, and sea ice), and their interactions. They are run as computer simulations, to predict climate variables such as temperature, pressure, precipitation etc. over multiple centuries. Several GCMs have been proposed by climate science institutes
from different countries around the world. 
The forecasts of future climate variables as predicted by these models have high variability which in turn introduces uncertainty in analysis based on these predictions \cite{IPCC5Report}. The main reason for such uncertainty in the response of the GCMs is due to model variability and can be greatly reduced by suitably combining outputs from multiple GCMs \cite{IPCC5Report}. 


In this analysis we consider the problem of GCM outputs combination for land surface
temperature prediction in South America. Being the world's fourth-largest continent, covering
approximately 12\% of the Earth's land area, the climate of South America varies considerably. The Amazon river
basin in the north has the typical hot wet climate suitable for the growth of rain forests. The Andes
Mountains, on the other hand, remain cold throughout the year. The desert regions of Chile is the
driest part of South America.

We use 10 GCMs from the CMIP5 dataset~\cite{tasm12}.
The details about the origin and location of the datasets are listed in Table~\ref{table:GCM_info}.

The global observation data for surface temperature is obtained from the Climate Research Unit
(CRU)\footnote{http://www.cru.uea.ac.uk}. We align the data from the GCMs and CRU
observations to have the same spatial and temporal resolution,
using publicly available climate data operators (CDO)\footnote{https://code.zmaw.de/projects/cdo}.
For all the experiments, we used a $2.5^o \times 2.5^o$ grid over latitudes and longitudes in South America,
and monthly mean temperature data for 100 years, 1901-2000, with records starting from January 16, 1901.
In total, we consider 1200 time points (monthly data) and 250 spatial locations over land\footnote{Dataset is available
at first author's home page.}. For the MTL
framework, each geographical location forms a task (regression problem).\\

\noindent{\textbf{Baselines and Evaluation:}} We consider the following 4 baselines for comparison and
evaluation of MSSL performance. We will refer to these baselines and MSSL as the ``models'' in
the sequel and the constituent GCMs as ``submodels''. The four baselines are:

\begin{enumerate}
\item {\bf Average Model} is the current technique used by Intergovernmental
Panel on Climate Change (IPCC)\footnote{http://www.ipcc.ch}, which gives equal weight to all
GCMs at every location;

\item {\bf Best GCM} which uses the predicted outputs of the best
GCM in the training phase (lowest RMSE), this baseline is not a combination of models, but a single GCM instead;

\item {\bf Linear Regression} is an Ordinary Least Squares (OLS) regression for each geographic location;

\item {\bf Multi Model Regression with Spatial Smoothing  (S$^2$M$^2$R)} is the model recently
proposed by~\cite{Karthik2013}, which is a special case of MSSL with pre-defined dependence matrix
$\boldsymbol{\Omega}$ equal to the Laplacian matrix. This model is referred to as S$^2$M$^2$R.
It incorporates spatial smoothing using the graph Laplacian over a grid graph.
\end{enumerate}

For our experiments, we considered a moving window of 50 years of data for training and the next
10 years for testing. This was done over 100 years of data, resuting in 5 train/test sets.
Therefore, the results are reported as the average RMSE over these test sets.

Table~\ref{tab:perf_climate} reports the average and standard deviation RMSE for all 250 geographical
locations. While average model has the highest RMSE, $r$-MSSL has the smallest RMSE in comparison
to the baselines. The performance of OLS and S$^2$M$^2$R are very similar.

\begin{table}[htb]
\centering
\begin{tabular}{ccccc} \hline
               {\bf Average}    & {\bf Best GCM}         &  {\bf OLS}      &  {\bf S$^2$M$^2$R}  &   {\bf r-MSSL}    \\ \hline
                1.621        & 1.410                  & 0.866           &  0.863                  &   \bf{0.780}   \\
				($\pm$0.020)  & ($\pm$0.037)          & ($\pm$0.037)    &  ($\pm$0.067)         &    ($\pm$0.039)   \\
              \hline
\end{tabular}
\caption{Mean and standard deviation of RMSE over all locations for $r$-MSSL and the baseline algorithms.
$r$-MSSL performs best in predicting temperature over South America.}
\label{tab:perf_climate}
\end{table}

\begin{figure}[htb]
\centering
\includegraphics[scale=0.6]{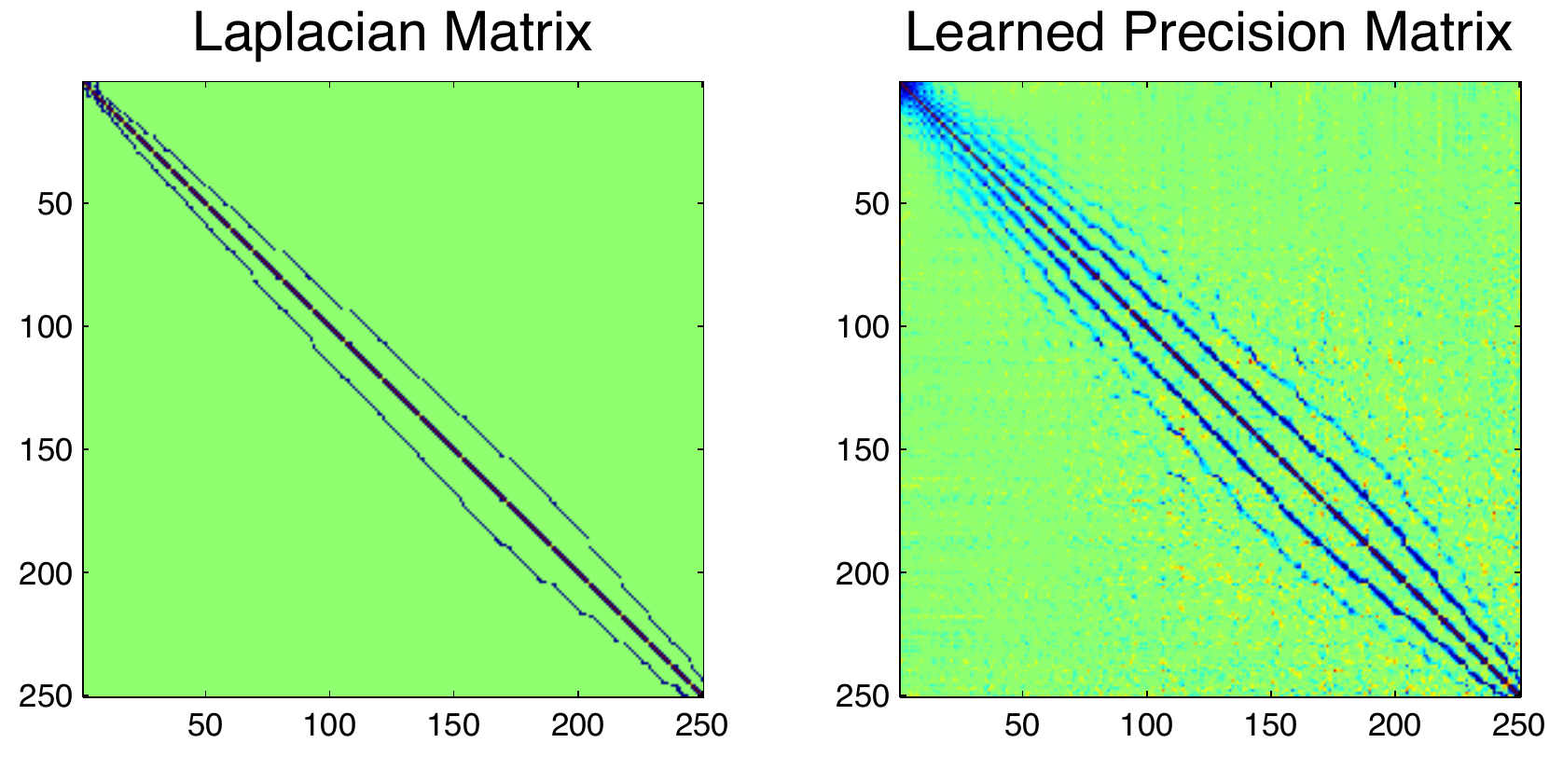}
\caption{The Laplacian matrix (on grid graph) assumed by S$^2$M$^2$R (left) and the Precision matrix learned by $r$-MSSL (right). $r$-MSSL can capture spatial relations beyond immediate neighbors.}
\label{fig:prec_mat_res}
\end{figure}

Figure~\ref{fig:prec_mat_res} shows the precision matrix estimated by $r$-MSSL algorithm and the Laplacian matrix assumed by
S$^2$M$^2$R. Not only is the precision matrix for $r$-MSSL, able to capture the relationship
between a geographical locations' immediate neighbors (as in a grid graph) but it also recovers
relationships between locations that are not immediate neighbors.



\begin{figure}[htb]
\centering
\includegraphics[scale=0.7]{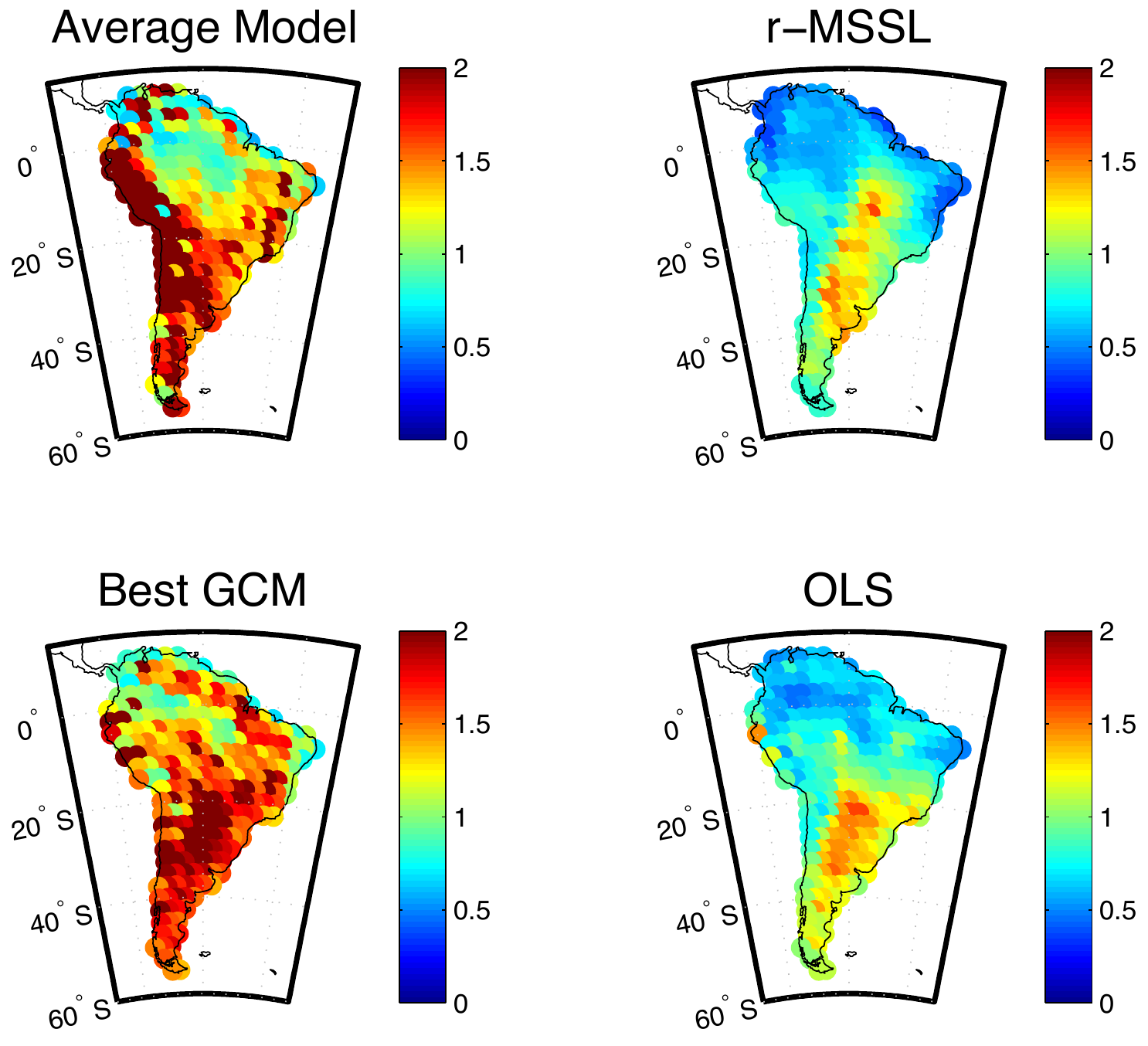}
\caption{[Best viewed in color] RMSE per location for each one of the four approaches. Since
S$^2$M$^2$R produced almost the same RMSE than OLS, it is not shown in the figure.}
\label{fig:error_plot}
\end{figure}

The RMSE per geographical location for average model and $r$-MSSL is shown in Figure~\ref{fig:error_plot}.
As previously mentioned, South America has diverse climate and not all of the GCMs are designed
to take into account and capture this. Hence, averaging the model outputs as done by IPCC, reduces
prediction accuracy. On the other hand $r$-MSSL performs better because it learns the right weight
combination on the model outputs and incorporates spatial smoothing by learning the task relatedness.


\begin{figure}[htb]
\centering
\includegraphics[scale=0.6]{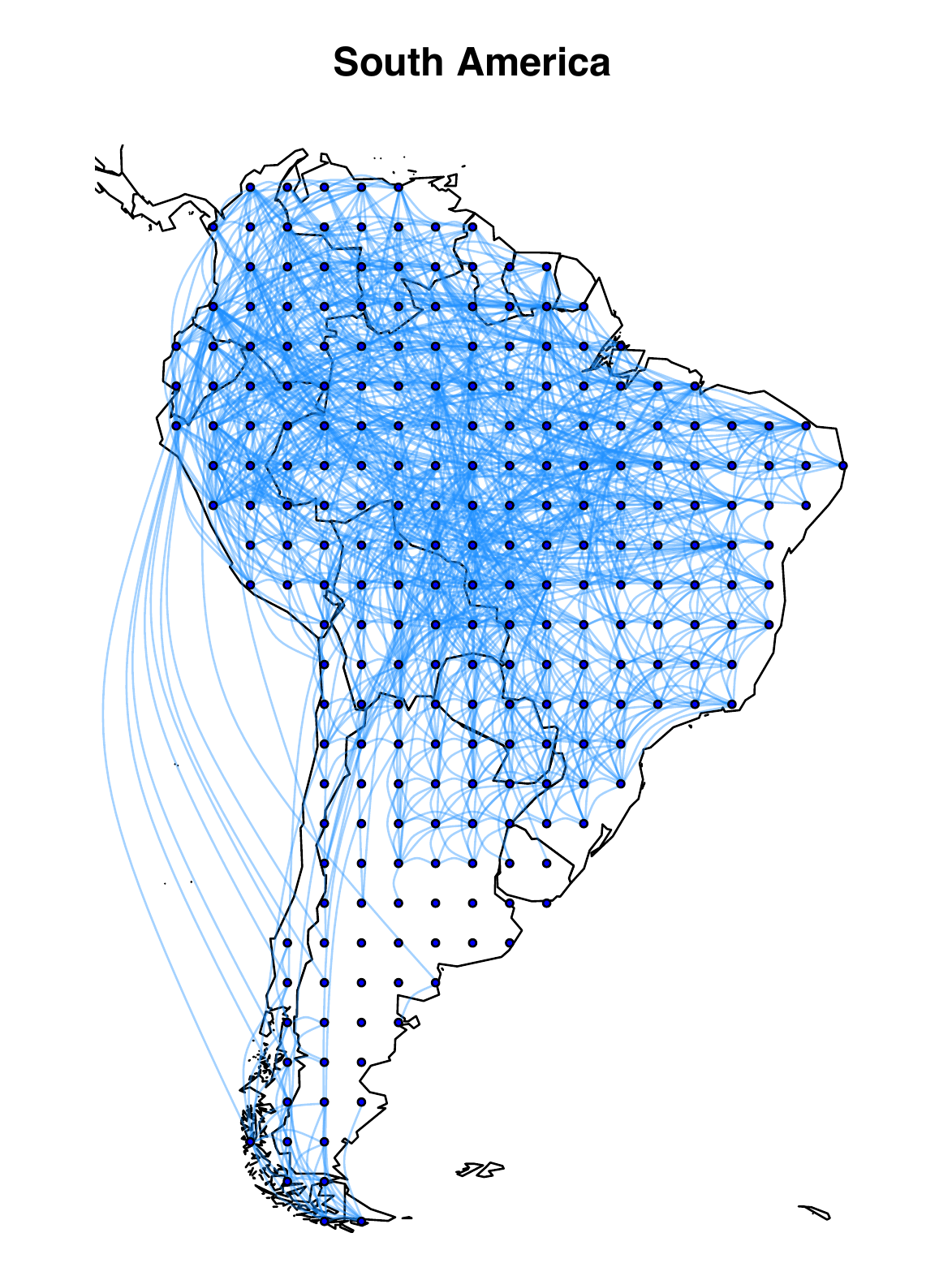}
\caption{Relationships between geographical locations estimated by $r$-MSSL algorithm. The blue
lines indicate that connected locations are dependent on each other. Also see Figure~\ref{fig:chord_mat_res}
for a chord diagram of the connectivity graph.}
\label{fig:map_connections}
\end{figure}

Figure~\ref{fig:map_connections} presents the relatedness structure estimated by $r$-MSSL among the geographical
locations. The regions connected by blue lines are dependent on each other. We immediately observe that locations in the northwest part of
South America are densely connected. This area has a typical tropical climate and comprises the Amazon
rainforest which is known for having hot and humid climate throughout the year with low temperature variation \cite{Ramos2014}.


The cold climates which occur in the southernmost parts of Argentina and Chile are clearly highlighted.
Such areas have low temperatures throughout the year, but there are large daily
variations \cite{Ramos2014}.

An important observation can be made about South America west cost, ranging from central Chile to
Venezuela passing through Peru which has one of the driest deserts in the world. These areas are located
to the left side of Andes Mountains and are known for arid climate. The average model
is not performing well on this region compared to $r$-MSSL. We can see the long lines connecting these
coastal regions, which probably explains the improvement in terms of RMSE reduction achieved by $r$-MSSL.
The algorithm uses information from related locations to enhance its performance on these areas.

The lack of connecting lines in central Argentina can be explained by the temperate climate which
presents greater range of temperatures than the tropical climates and may have extreme climatic
variations. That is a transition area between the very cold southernmost region and the hot
and humid central area of South America. It also comprises Patagonia, a semiarid scrub plateau that
covers nearly all of the southern portion of mainland, whose climate is strongly influenced by
the South Pacific air current which increases the regions temperature variability \cite{Ramos2014}.
Due to such high variability it becomes harder to provide accurate temperature predictions.

\begin{figure}[htb]
\centering
\includegraphics[scale=0.45]{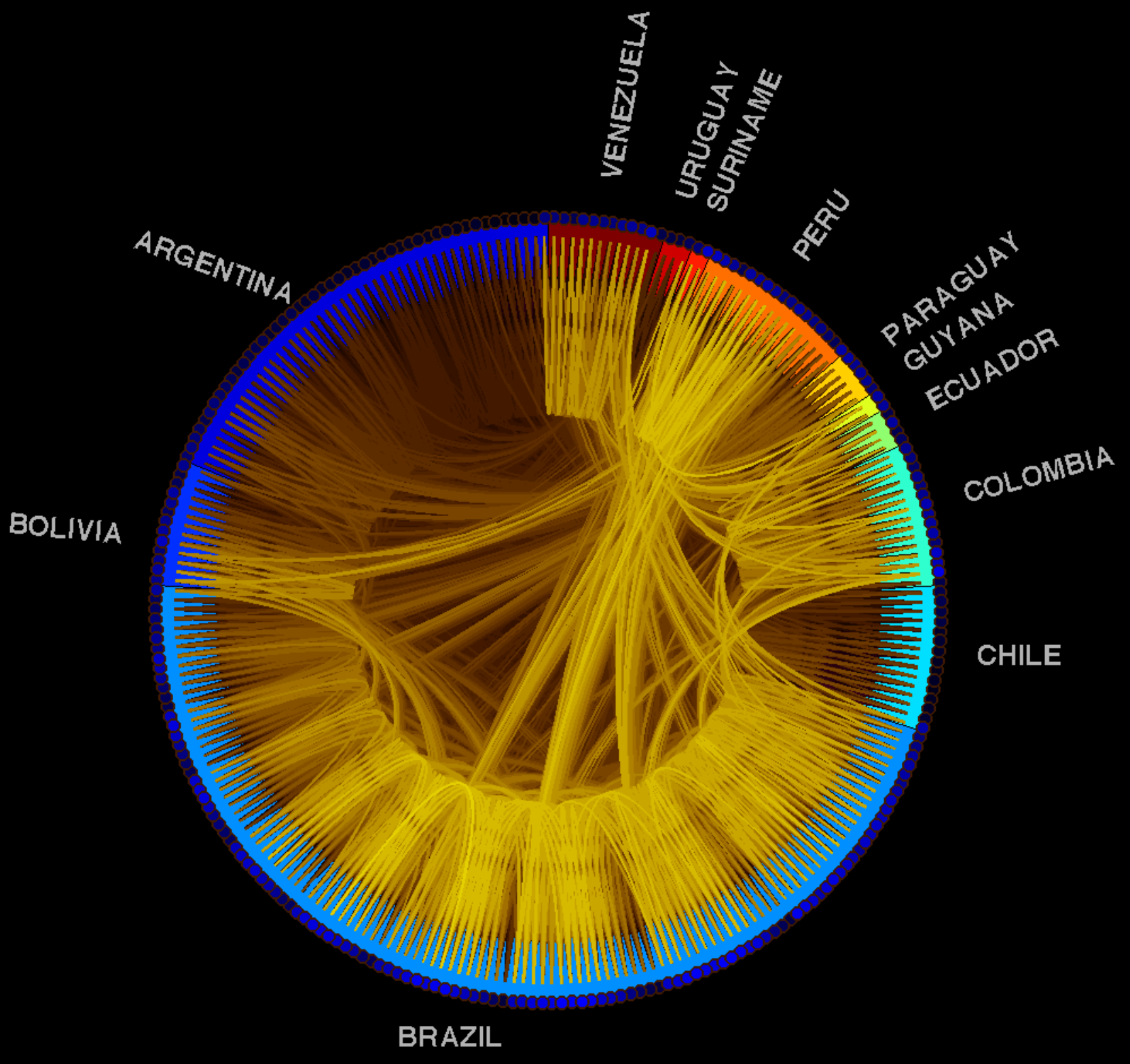}
\caption{[Best viewed in color] Chord graph representing the structure estimated by $r$-MSSL algorithm.}
\label{fig:chord_mat_res}
\end{figure}

Figure~\ref{fig:chord_mat_res} presents the dependency structure using a chord diagram.
Each point on the periphery of the circle is a location in South America and represents the task of
learning to predict temperature at that location. The locations are arranged serially on the periphery
according to the respective countries. We immediately observe that the locations in Brazil are heavily
connected to parts of Peru, Colombia and parts of Bolivia. These connections are interesting as these
parts of South America comprise the Amazon rainforest. We also observe that locations within Chile and
Argentina are less densely connected to other parts of South America. A possible explanation could be
that while Chile which includes the Atacama Desert is a dry region located to the west of the Andes,
Argentina, especially the southern part experiences heavy snowfall which is different from the hot and
humid rain forests or the dry and arid deserts on the west coast. Both these regions experience climatic
conditions which are disparate from the northern rain forests and from each other. The task dependencies
estimated from the data reflect this disparity.

\subsection{Classification}


For the classification problems we present experimental results on four well known classification
benchmark datasets.

\noindent{\bf Datasets:} We test the performance of the proposed MSSL algorithm on the following four data sets:

\noindent{\bf (a) Landmine Detection:} Data from 19 different landmine fields were collected, which have distinct types of
characteristics. Each object in a given data set is represented by a 9-dimensional feature vector and the corresponding
binary label (1 for landmine and 0 for clutter) \cite{Xue2007}.
The feature vectors are extracted from radar images, concatenating four moment-based features, three
correlation-based features, one energy ratio feature and one spatial variance feature. The goal is to classify between mine or clutter.

\noindent{\bf (b) Spam Detection:} Email spam dataset from ECML 2006 discovery challenge\footnote{http://www.ecmlpkdd2006.org/challenge.html}.
This dataset consists of two problems: In Problem A, we have emails from 3 different users (2500 emails per user);
whereas in Problem B, we have emails from 15 distinct users (400 emails per user). We performed feature selection
to get the 500 most informative variables using Laplacian Score feature selection algorithm \cite{He:2006}. The
goal is to classify between spam vs. ham. For both problems, we create different tasks for different users.

\noindent {\bf (c) Mnist:} MNIST dataset\footnote{http://yann.lecun.com/exdb/mnist/} consists of 28$\times$28-size images
of hand-written digits from 0 through 9. We create binary classification problems for each pair of digits,
totaling 45 tasks. The number of samples for each classification problem is about 15000.

\noindent {\bf (d) Letter:} The handwritten letter dataset\footnote{http://ai.stanford.edu/$\sim$btaskar/ocr/} consists
of eight tasks each of which is a binary classification problem for two letters: a/g, a/o, c/e, f/t, g/y, h/n, m/n and i/j.
The input for each data point consists of 128 features representing the pixel values of the handwritten letter.
The number of data points for each task varies from 3057 to 7931.\\


\noindent{\bf Baseline algorithms:} Four baselines algorithms were considered in the experiments and
the regularization parameters for all algorithms were selected using cross-validation from \{0.01, 0.1, 1, 10, 100\}.
The algorithms are:

\begin{enumerate}
\item {\bf Logistic Regression (LR)}, learns separate logistic regression models for each task;

\item {\bf Joint Feature Selection (JFS)} \cite{Argyriou:2007} employs a
$\ell_{2,1}$-norm regularization term to capture the task relatedness
from multiple related tasks constraining all models to share a common set
of features;

\item {\bf CMTL} \cite{zhou2011a} incorporates a regularization term
to induce clustering between tasks and then share information only to tasks
belonging to the same cluster;

\item {\bf Low rank MTL} algorithm \cite{Abernethy:2006}
assumes that related tasks share a low dimensional subspace and applies a
trace regularization norm to capture that.
\end{enumerate}

\noindent{\bf Results:} Table~\ref{tab:perf} shows the results obtained by each algorithm for all datasets.
\begin{table*}[htb]
\centering
\begin{tabular}{lccccc}
\hline
     & \multicolumn{5}{c}{\bf{Algorithms}} \\ \cline{2-6}
                   & LR                 &        CMTL           &       Low Rank         &        JFS   & $p$-MSSL\\
\hline
    Landmine       & 5.82($\pm$9e-2)    &  5.85($\pm$2e-2)    & 5.74($\pm$2e-2)        & 5.92($\pm$2e-2)         &  \bf{5.55($\pm$\bf{4e-2})}\\
    Spam-3 users   & 3.48($\pm$2e-1)    &  3.29($\pm$4e-1)    & 2.47($\pm$6e-2)        & 7.81($\pm$1e-1)         & \bf{1.84($\pm$8e-2)}  \\
    Spam-15 users  & 13.50($\pm$9.5e-1) &  9.99($\pm$1e-1)    & \bf{7.04($\pm$5.6e-1)} & 11.86($\pm$2.4e-1)      & \bf{6.84($\pm$1.3e-1)}\\
    Mnist (digit)  & 2.55($\pm$3.8e-1)  &  2.86($\pm$3e-2)    & 2.16($\pm$2.6e-2)      & 4.01($\pm$1.7e-2)       & \bf{2.05($\pm$1.5e-2)} \\
    Letter         & 5.24($\pm$4.6e-2)  &  5.12($\pm$1.8e-2)  & 7.41($\pm$6.6e-2)      & \bf{5.10($\pm$4.8e-2)}  &  \bf{5.08($\pm$3.6e-2)}  \\
 \hline
\end{tabular}
\caption{Average classification error rates and standard deviation (10 independent runs) of different methods for all five considered datasets.
Bold values indicate a significant statistical improvement in relation to others methods at $\alpha=0.05$.}
\label{tab:perf}
\end{table*}

\begin{figure}[htb]
	\centering
    \includegraphics[scale=0.35]{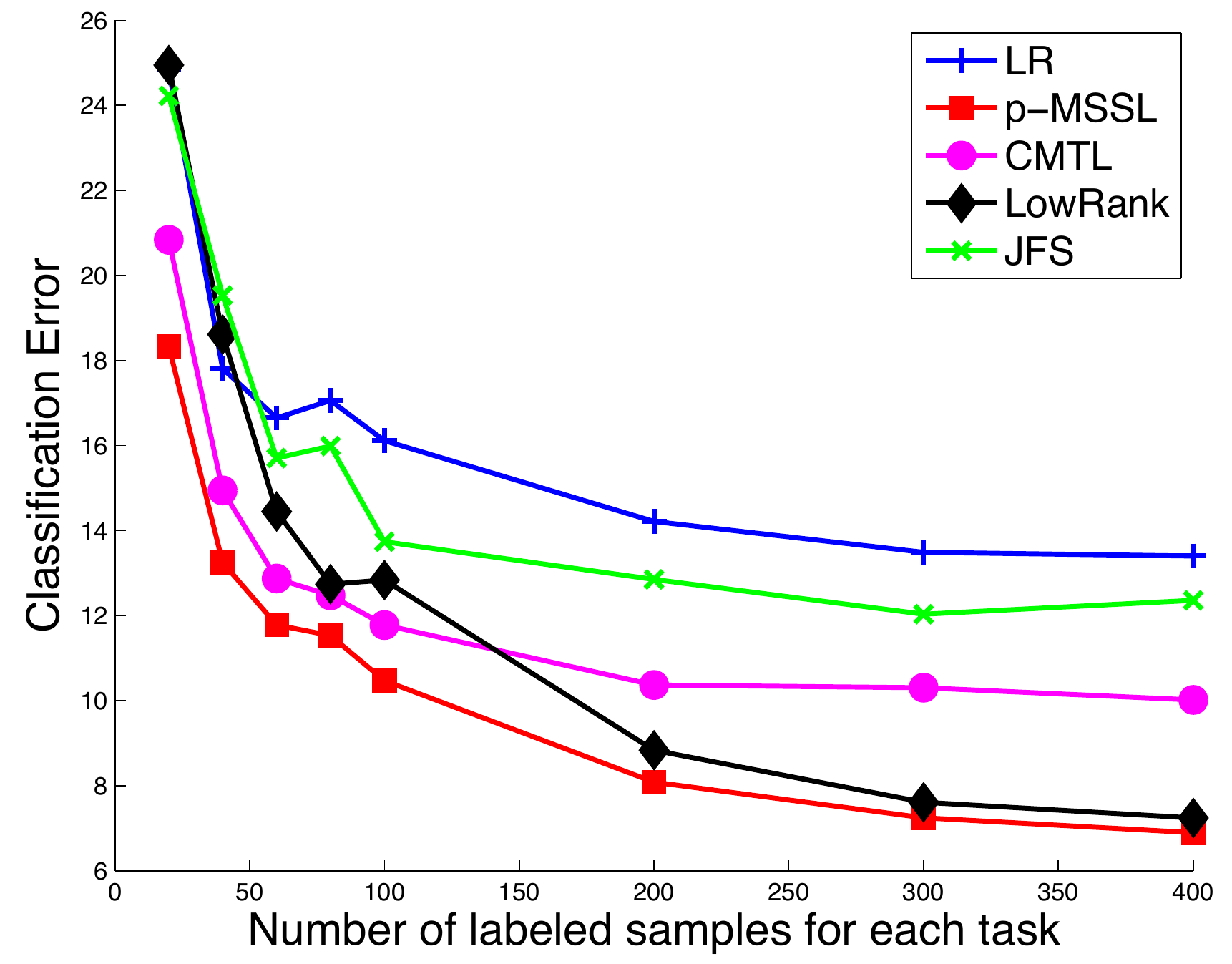}
    \caption{Average classification error obtained from 10 independent runs versus number of training
	data points for all tested methods on \textit{Spam-15-users} dataset.}
    \label{fig:varsize}
\end{figure}

Figure~\ref{fig:varsize} shows the behavior of each algorithms when the number of labeled samples for each task varies. MTL algorithms have better performance compared to LR when there are few labeled samples available. $p$-MSSL also gives better results for all
ranges of sample size when compared to the other algorithms.

\begin{figure}[htb]
    \centering
	\includegraphics[scale=0.5]{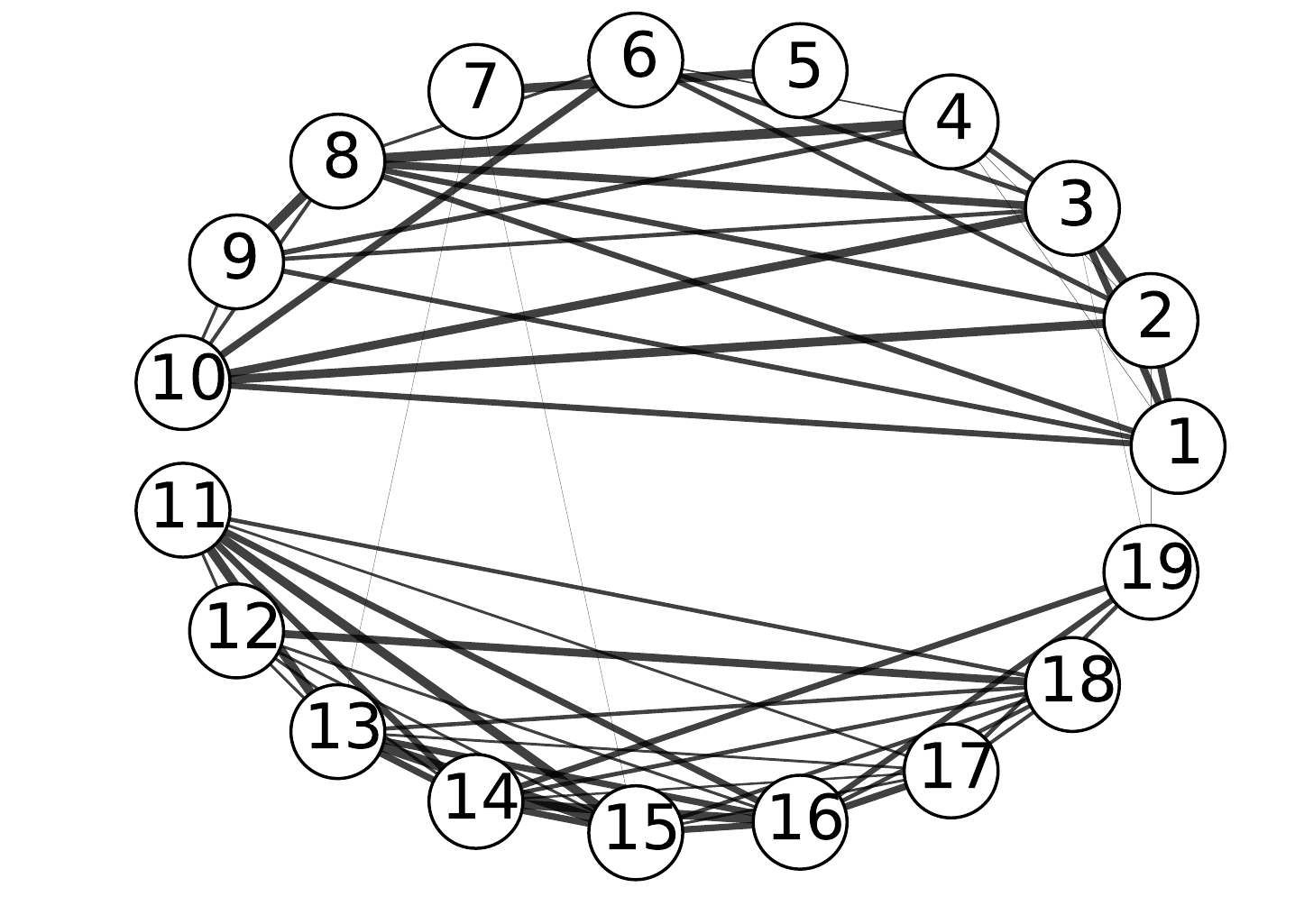}
    \caption{Graph representing the dependency structure among tasks captured by precision matrix estimated by $p$-MSSL.
	Tasks from 1 to 10 and from 11 to 19	are more densely connected to each other, indicating two clusters of tasks.}
    \label{fig:corr_landmine}
\end{figure}

In the Landmine detection dataset, samples from tasks 1-10 were collected at foliated regions and 11-19 are collected
at regions that are bare earth or desert (these demarcations are good, but not absolutely precise, since some barren
areas have foliage, and some largely foliated areas have bare soil as well).
Therefore we expect two dominant clusters of tasks, corresponding to the two different types of ground surface
conditions. In Figure~\ref{fig:corr_landmine} we show the graph structure representing the precision matrix
estimated by $p$-MSSL. One can see that tasks from foliate regions (1-10) are densely connected to each other
while tasks with data input from desert areas (11-19) also form a cluster.

\section{Conclusion}
\label{sec:conclusion}

In this paper we propose a framework for multi-task structure learning. Our proposed framework
simultaneously learns the tasks and their relatedness, with the task dependencies defined as edges
in an undirected graphical model. The problem formulation leads to a bi-convex problem which can be efficiently
solved using alternating minimization. We show that the proposed framework is general enough to be
specialized to Gaussian models and GLM's. Extensive experiments on benchmark and climate datasets
illustrate that structure learning not only improves multi-task prediction performance, but also
captures very informative correlated behaviors within tasks.

\noindent {\bf Acknowledgments:}
The research was supported by NSF grants IIS-1029711, IIS-0916750,
IIS-0953274, CNS-1314560, IIS-1422557, CCF-1451986, IIS-1447566,
and by NASA grant NNX12AQ39A. Arindam Banerjee acknowledges support from
IBM and Yahoo. Fernando J Von Zuben would like to thank CNPq for the financial support. Andr\'e R Gon\c{c}alves was supported by Science
without Borders grant from CNPq, Brazil. Access to computing facilities were provided by University of
Minnesota Supercomputing Institute (MSI).

\bibliographystyle{natbib}
\bibliography{references}

\end{document}